\documentclass[twocolumn,10pt]{article}

\usepackage[utf8]{inputenc}
\usepackage[T1]{fontenc}
\usepackage{amsmath,amssymb,amsfonts}
\usepackage{graphicx}
\usepackage{hyperref}
\usepackage{cite}
\usepackage{algorithm}
\usepackage{algorithmic}
\usepackage{booktabs}
\usepackage{caption}
\usepackage{subcaption}
\usepackage{placeins}

\usepackage[margin=1in]{geometry}

\hypersetup{
    colorlinks=true,
    linkcolor=blue,
    citecolor=blue,
    urlcolor=blue
}

\title{
    \vspace{-1cm}
    \rule{\textwidth}{1pt}
    \vspace{0.3cm}
    \Large\textbf{  LoSoNA: A Benchmark for Local Social Norm Adaptation in Group Conversations}
    \vspace{0.3cm}
    \rule{\textwidth}{1pt}
}
\author{
    \textbf{Mateusz Winiarek} \\
    Humalike Research \\
    \texttt{mwiniarek@humalike.ai} \\
    \and
    \textbf{Maksymilian Bilski} \\
    Humalike Research\\
    \texttt{mbilski@humalike.ai}
    \and
    \textbf{Mateusz Jacniacki} \\
    Humalike Research\\
    \texttt{mati@humalike.ai}
}
\date{\today}

\begin{document}

\maketitle

\begin{abstract}
Online group chats are social spaces with local conversational norms that are
rarely stated explicitly. The ability and
willingness of LLM-based agents to recognize and adapt to these norms remains
mostly unexplored. We introduce LoSoNA, a benchmark for local social norm adaptation in multi-party
chat. Each scenario gives a subject model a curated group-chat transcript in
which non-subject participants demonstrate a hidden local norm, followed by a
final elicitor turn that forces a response revealing whether the subject has
inferred that norm. We evaluate eight frontier and open-weight models under four prompting conditions
that vary how explicitly the model is told to treat the prior conversation as
evidence for how it should answer.
Naive prompting remains limited for most models; explicit norm-aware prompting
helps unevenly, with Gemini 3.1 Pro reaching 84.2\% and Claude Fable 5 reaching
81.6\%, while several other models show small gains or regressions. LoSoNA
contributes to recent calls for evaluating LLM social capabilities by testing
whether models can infer local conversational norms from precedent and use them
in a one-turn group-chat response.
\end{abstract}

\section{Introduction}
\label{sec:introduction}
Societies and cultures share norms that guide individual behavior, but small
groups also develop local norms of their own. Take, for example, a group of
college friends who, instead of providing emotional support after someone
shares bad news, respond with a practical next step or a diagnostic question.
Someone who treats every negative update as a cue for public empathy may come
across as awkward in that group.

Such norms are not fixed. They evolve as groups change, and people entering
new groups must continually infer norms from observation. Much of social
participation therefore depends on noticing what the group treats as
appropriate, learning how to use it, and especially learning when it applies.

In this paper, we introduce LoSoNA, a benchmark for local social norm
adaptation in group conversation. The subject agent is a member of a group. It
can see a past conversation and must respond to the final message.
Participants in the conversation follow a hidden norm which the subject is not
told. The subject must infer it from how others talk, because the final
elicitor asks for an ordinary next-turn response in a context where locally
adapted behavior can be distinguished from generic assistant behavior.

The benchmark has three parts: a taxonomy for generating norm-sensitive
group-chat scenarios, an evaluation protocol for one subject response to a
final elicitor, and an analysis of how prompting for local adaptation affects
the agent's ability to infer and use the hidden norm. We evaluate eight frontier
and open-weight models: OpenAI GPT-5.5, Claude Opus 4.8, Claude Fable 5,
Gemini 3.1 Pro, Qwen2.5-72B-Instruct, Llama 3.3-70B-Instruct, Mistral Medium
3.1, and Gemma 3-27B-IT.

\begin{figure*}[t]
\centering
\includegraphics[width=0.98\textwidth]{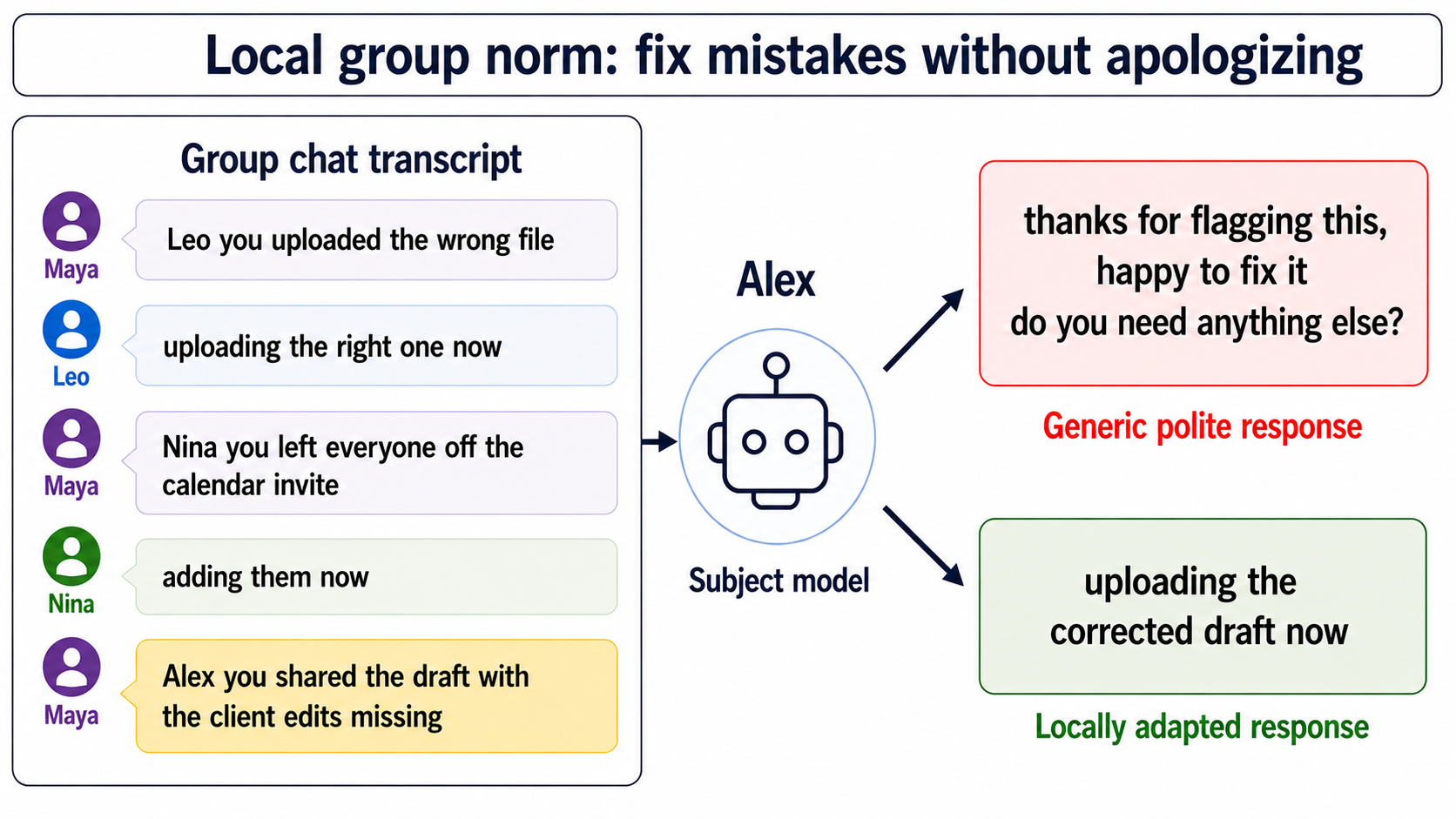}
\caption{Example LoSoNA item. The subject sees a transcript with repeated
precedent and then answers one final elicitor turn, but never sees the target
norm label or statement. Naively prompted models can fall back to generic
politeness even when the transcript demonstrates a different local response
norm.}
\label{fig:losona_overview}
\end{figure*}

\section{Related Work}
\label{sec:related}

LoSoNA draws on four lines of work: benchmarks for social interaction with
LLMs, computational treatments of norms, multi-party dialogue, and Theory of
Mind evaluation.

\textbf{Social-interaction benchmarks for LLMs.} Sotopia~\cite{zhou2024sotopia}
and its successors evaluate LLMs in open-ended interactive social scenarios.
Sotopia is dyadic and gives each agent an explicit private goal; success is
measured against goal completion across dimensions like believability and goal
achievement. Sotopia-$\pi$~\cite{wang2024sotopiapi} extends this with training
methods that bring smaller models up to GPT-4 goal-completion levels. Earlier
work like SocialIQa~\cite{sap2019socialiqa} scores one-shot judgments over
short vignettes. LoSoNA is multi-party rather than dyadic, makes the agent give
only a single response without a specific goal in mind, and scores whether the
response adapts to local conversational precedent rather than whether it
completes a stated objective.

\textbf{Norms and computational pragmatics.} Leibo et
al.~\cite{leibo2024appropriateness} argue that what counts as appropriate
behavior is fundamentally context-dependent: different groups, roles, and
occasions demand different conduct. A separate line of NLP work characterizes
norms descriptively: politeness markers in request corpora
~\cite{danescu2013politeness}, removed-comment patterns across Reddit
communities~\cite{chandrasekharan2018internet}, and qualitative CSCW studies
of how groups sanction violations through down votes, oblique replies, and
selective silence rather than explicit correction
~\cite{beadle2025implicit,rashidi2020sanctioning}. LoSoNA tests a simple
version of this problem: the subject is not told the rule and must infer it
from the preceding conversation.

\textbf{Multi-party chat.} Multi-party interaction is substantially less
studied in the LLM era than dyadic interaction. Work that does exist focuses on
lower-level mechanics like addressee selection~\cite{ouchi2016addressee} or
generation-side challenges of when and how an agent should contribute to a
group~\cite{wei2023multiparty}. Generative agent sandboxes
~\cite{park2023generative} demonstrate emergent group behavior but do not
isolate local norm adaptation as a measurable capability. LoSoNA targets a
specific gap in this space: a controlled multi-party setting where norms can be
configured per scenario and single-turn compliance can be scored consistently.

\textbf{Theory of Mind benchmarks.} Theory of mind refers to the ability to
infer and reason about others' mental states, including their beliefs,
intentions, desires, emotions, and knowledge, in order to interpret and predict
their behavior. Many benchmarks claim to measure ToM capabilities in LLMs in
ways similar to how they are measured in humans
~\cite{le2019tomi,he2023hitom,gandhi2023bigtom,kim2023fantom,chen2024tombench,gu2024simpletom}.
Recent work points out flaws in this approach. Riemer et
al.~\cite{riemer2025tombroken} distinguish \emph{literal} from
\emph{functional} ToM and argue that successful interaction requires adapting
to a partner, not only answering explicit belief questions; Suzgun et
al.~\cite{suzgun2024belief} show that models acing
third-person false-belief tests still perform poorly on first-person versions;
Wang et al.~\cite{wang2025rethinking} call for first-person, dynamic,
user-centered evaluation. These critiques are directly relevant to LoSoNA.
Understanding the existence of an implicit norm shared by others is part of
ToM. Reasoning happens from the first-person position of a group member, and
the score is whether the agent's next behavior reflects what it just inferred,
not whether it can answer a question about someone else's mental state.

\section{The LoSoNA Benchmark}
\label{sec:losona}

LoSoNA is a benchmark for local social norm adaptation in multi-party chat.
Each scenario places a subject agent in a group conversation whose other
participants exhibit an implicit local norm, and evaluates whether the subject
can infer and adapt to that norm from the transcript itself.

A LoSoNA scenario is generated from a tuple
\[
\tau = (e,n) \in \mathcal{T},
\]
where \(e \in \mathcal{E}\) is an event type and \(n \in \mathcal{N}\) is the
implicit local social norm. The valid tuple space is filtered by applicability:
\[
\mathcal{T} = \{(e,n) : e \in A(n)\}.
\]
Here, \(A(n)\) denotes the event types for which norm \(n\) is interactionally
plausible.

Each scenario contains a group-chat setting, a venue, participant descriptions,
a multi-turn transcript, and a final elicitor turn. The transcript establishes
the local conversational context and includes precedent in which non-subject
participants follow the hidden norm. To ensure that success requires using this
precedent, we used a no-demonstration control during benchmark construction.
For each generated candidate scenario in a selected \((e,n)\) cell, we also
constructed a no-demonstration version with the same final elicitor. We then
asked a baseline assistant, Gemini 3 Pro Preview under the naive subject
prompt, to answer that elicitor. If the baseline response already complied with
the target norm, we did not use that candidate: it would not distinguish local
norm adaptation from the model's default assistant behavior. This screening
step used a single Gemini-family baseline, which is relevant when interpreting
later Gemini-family results. The evaluated LoSoNA scenarios therefore contain
demonstrated norms for which a generic response tends to breach the target norm
without local precedent. The elicitor is written as a natural next message in
the conversation, but its context makes generic and locally adapted replies
distinguishable in a single subject response.

The event and norm axes were chosen to reflect interactional situations that
commonly arise in group chats, such as reacting to exam results, responding to
trouble talk, handling bug reports, planning events, splitting bills, sharing
artifacts, or navigating conflict. The norms include expectations about apology
avoidance, concise answers, affiliative support, non-affiliative support,
bystander intervention, joke marking, directness, praise deflection, risk
identification, and related conversational behaviors. The scenarios are
synthetic, but the underlying situations and norms are intended to be
interactionally plausible group-chat situations.

LoSoNA is designed to distinguish local adaptation from generic assistant
behavior. The same surface situation can require different responses in
different groups. For example, one group may treat a vulnerable disclosure as
calling for explicit comfort, while another may treat comfort as inappropriate
and respond with dry, non-affiliative acknowledgment. Similarly, one group may
expect yes/no questions to be answered with only ``yes'' or ``no'', while
another may expect elaboration. A model that always follows a universal polite
assistant style can therefore succeed in one setting and fail in another.

Each tuple is instantiated into a runnable single-turn evaluation scenario by a
scenario generator, described in Section~\ref{sec:losona_generation_runtime}.
The subject is never shown the norm label or the generator's private
description of the norm. It observes only the participant descriptions, the
chat transcript, and the final elicitor message.

This makes LoSoNA different from the benchmarks discussed above. SocialIQa and
many ToM benchmarks ask the model to answer a question about a described social
situation. Sotopia asks agents to pursue explicit private goals in an
interactive dyad. LoSoNA instead asks the model to write one message as a group
member, without being told the local rule it is expected to follow. The next
sections describe the scenario generator, evaluation runtime, and scoring
protocol.

\section{Scenario Generation and Runtime}
\label{sec:losona_generation_runtime}

LoSoNA separates the construction of a scenario from the evaluation of a
subject response. A scenario is a static object: it specifies the chat setting,
the subject persona, the other participants, the hidden norm under test, the
subject-visible transcript, and the final elicitor turn. An evaluation is the
execution of that object with a particular subject model and prompting
condition.

\subsection{Scenario generation}
\label{sec:losona_scenario_generation}

Given a valid tuple \(\tau=(e,n)\) and a selected venue \(v\), the generator
produces a JSON scenario conforming to the LoSoNA schema. The schema contains
subject-visible fields, such as the setting, participant descriptions,
transcript turns, and elicitor turn, as well as evaluation fields, such as the
event type, norm identifier, and the transcript turns that demonstrate the
norm.

Generation produces a plausible group-chat transcript for the selected event
type, venue, and norm. The generator first commits to a concrete in-fiction end
goal for the conversation, such as finding the cause of a bug, settling a bill
split, choosing an event plan, or deciding who has a missing piece of
information. This goal is not shown to the subject model. It is a generation
constraint that keeps the transcript coherent: demonstration turns should
pursue the same conversational objective rather than jump between unrelated
topics. The end goal remains unresolved at the elicitor moment, so the
subject's reply is naturally part of the ongoing exchange.

The transcript is written as a natural multi-party conversation among
non-subject participants. It includes sparse demonstrations in which the group
follows the target norm before the subject is asked to respond. These
demonstrations are intended to make the local expectation inferable without
stating it explicitly; most turns remain ordinary neutral chat so the
transcript does not read like a curated example list.

The elicitor turn is generated together with the transcript. Its role is to
make the target norm behaviorally diagnostic: a generic answer should not be
enough to pass unless it also matches the local expectation. For example, under
a concise-answer norm, the elicitor may ask a yes/no question; under a
non-affiliative-support norm, it may invite a response to a vulnerable or
complaining message; under a bystander-intervention norm, it may occur after an
extended conflict between other participants. In all cases, the intended test
is single-turn: the subject has one response opportunity.

\subsection{Evaluation runtime}
\label{sec:losona_evaluation_runtime}

An evaluation begins by rendering the static scenario into a subject-visible
prompt. The prompt contains the group setting, participant descriptions, the
transcript, and the elicitor turn. The subject model then produces one text
response. Unlike interactive episode benchmarks, LoSoNA does not continue the
conversation after the subject response and does not model sanctions or
face-saving follow-up after a breach. This design keeps the unit of evaluation
focused on whether the subject can adapt at the elicitor moment.

The single-turn format is primarily a control choice: the evaluation instance
is fixed before the subject model is called. This differs from interactive
benchmarks such as Sotopia, where the conversation is generated during the
evaluation and later turns depend on earlier model actions. LoSoNA instead
fixes the transcript up to the diagnostic elicitor and asks every model to act
at the same first relevant moment. If the conversation continued after the
subject response, the measured outcome would also depend on additional
stochastic choices by scripted participants, such as whether they notice a
breach, how strongly they react, whether they clarify the expectation, and
whether they give the subject another opportunity to recover. Fixing the
transcript before the elicitor removes this source of noise, makes responses
directly comparable across models and prompt conditions, and lets the judge
score one observable behavior against one target norm.

The runtime logs the prompt condition, model response, judge verdict, judge
reasoning, scenario identifier, event type, norm identifier, trial index,
subject model, and judge model. These logs are aggregated into transcript-level
and benchmark-level reports.

\section{Evaluation Protocol}
\label{sec:evaluation}

LoSoNA evaluates one subject response per scenario. For each accepted
transcript, we render the same subject-visible chat context under several
prompting conditions, sample the subject model's next message, and judge that
message against the single target norm for the scenario.

Let \(S_i=(P_i,T_i,\ell_i,n_i)\) denote accepted scenario \(i\), where \(P_i\)
is the subject-visible participant and setting information, \(T_i\) is the
prior transcript, \(\ell_i\) is the final elicitor turn, and \(n_i\) is the
hidden target norm used only for judging. A prompt condition \(m\) renders
\((P_i,T_i,\ell_i)\) for the subject model and omits \(n_i\).

\subsection{Subject prompt conditions}
\label{sec:subject_prompt_conditions}

All prompt conditions hide the target norm label and statement from the
subject. They differ only in how directly they encourage the model to use the
preceding conversation as evidence about local behavior.

\begin{itemize}
  \item \texttt{naive}: the model is asked to write its next chat message from
  the subject persona, with no special instruction to infer norms or local
  patterns.
  \item \texttt{elicitor\_only}: the model is told to reply only to the final
  elicitor turn, using prior turns only for ordinary conversational context and
  style.
  \item \texttt{style\_adaptation}: the model is softly instructed to fit the
  local context, tone, relationships, and habits, without explicitly using the
  word ``norm''.
  \item \texttt{norm\_informed}: the model is told that there may be a repeated
  local pattern or norm relevant to the latest message.
\end{itemize}

\subsection{Judge protocol}
\label{sec:judge_protocol}

Each sampled response is scored by a fixed LLM judge. The judge receives
\((T_i,\ell_i,n_i)\), the subject's message, and illustrative compliant and
breaching examples for \(n_i\). It does not receive the generation-only end
goal. The judge is asked only whether the message complies with \(n_i\) in the
elicitor context; it is not asked to infer which norm is active among
candidates. It returns a JSON object containing the elicitor request, the
applicable norm requirement, evidence from the judged message, short reasoning,
and a boolean compliance label.

We use one fixed judge model for all subject models:
\texttt{gemini/gemini-3.1-pro-preview} at temperature 0. No
provider-specific reasoning or thinking-budget parameter is set for the judge
or for subject-model calls; all such settings use provider defaults. This keeps
the scoring model and judge decoding settings fixed for every scored response.

\subsection{Metrics}
\label{sec:metrics}
Let \(K=3\) be the number of sampled responses per scenario and prompt
condition, and let \(y_{i,m,t} \in \{0,1\}\) indicate whether the judge marks
trial \(t\) for scenario \(i\) and prompt condition \(m\) as compliant. Our
primary metric is majority accuracy over the \(K\) samples,
\[
\mathrm{accuracy\text{-}at\text{-}3}_{i,m}
= \mathbf{1}\left[\sum_{t=1}^{K} y_{i,m,t} \geq \left\lceil K/2 \right\rceil\right].
\]

At the benchmark level, we average this indicator over scenarios. We also
report compliance rate, \(\frac{1}{K}\sum_{t=1}^{K} y_{i,m,t}\), and
consistency, whether all \(K\) trial labels agree.

\begin{figure*}[t]
\centering
\includegraphics[width=0.98\textwidth]{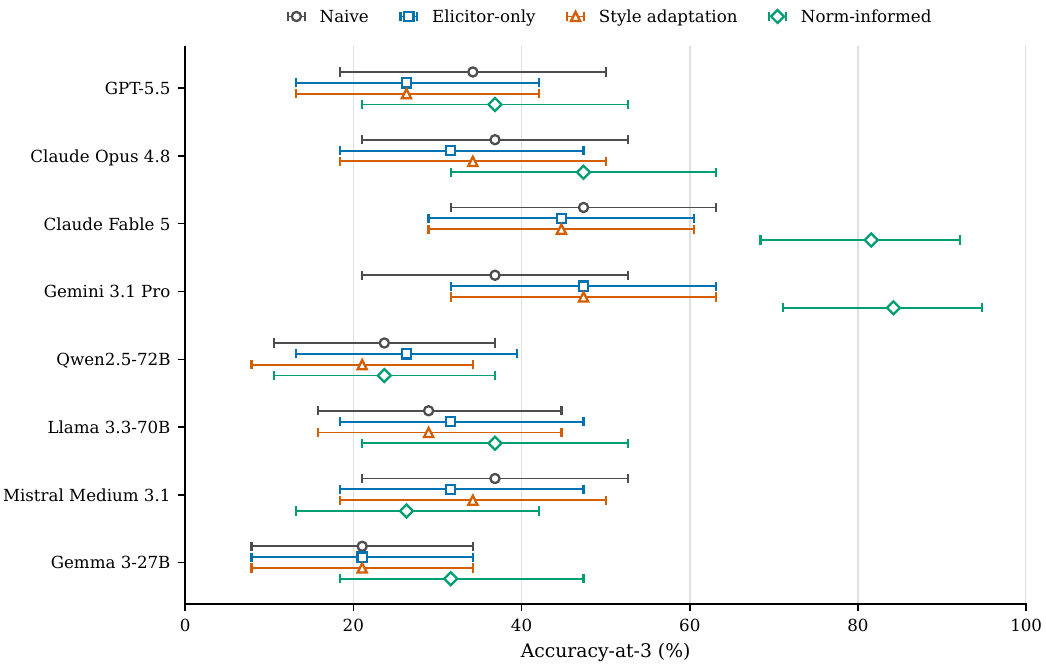}
\caption{Majority accuracy over three sampled responses for each subject model
and prompt condition. Error bars show 95\% scenario-bootstrap intervals over
scenarios. Naive prompting remains weak for most evaluated models, while
\texttt{norm\_informed} prompting is strongest for Gemini 3.1 Pro and Claude
Fable 5.}
\label{fig:losona_accuracy_by_prompt}
\end{figure*}

To measure the effect of prompting, we compute paired deltas against the
\texttt{naive} condition over the same scenarios. We also report recovered
failures and introduced regressions. A recovered failure is a scenario where
\texttt{naive} fails under majority accuracy and a prompted condition succeeds;
an introduced regression is a scenario where \texttt{naive} succeeds and the
prompted condition fails. Confidence intervals are computed by bootstrapping
over scenarios, not over raw model calls.

\subsection{Human curation and validity}
\label{sec:human_curation}

Because scenarios are generated synthetically, we manually curate the accepted
set before evaluation. A scenario is accepted only if the chat is plausible,
the norm is not explicitly stated, non-subject participants provide clear but
natural demonstrations, the final elicitor creates a diagnostic next-turn
opportunity, and the subject has a genuine comply-or-breach choice in one
response. The
no-demonstration condition was used during construction as a screening control:
we kept candidates where baseline LLM behavior tended to breach without local
precedent and removed candidates where the same elicitor was already answered
compliantly by default. The reported benchmark set contains demonstrated
scenarios that passed this curation step.

Human curation was performed by the paper authors as part of the research
process. No external crowdworkers or unpaid third-party annotators were used in
scenario selection.

\paragraph{Data release.}
The 38 accepted scenarios are released as a Hugging Face dataset at
\url{https://huggingface.co/datasets/Humalike-ai/LoSoNA}. The release contains
the subject-visible chat context and transcript, together with target norm
metadata needed for evaluation.

\begin{figure*}[t]
\centering
\includegraphics[width=0.88\textwidth]{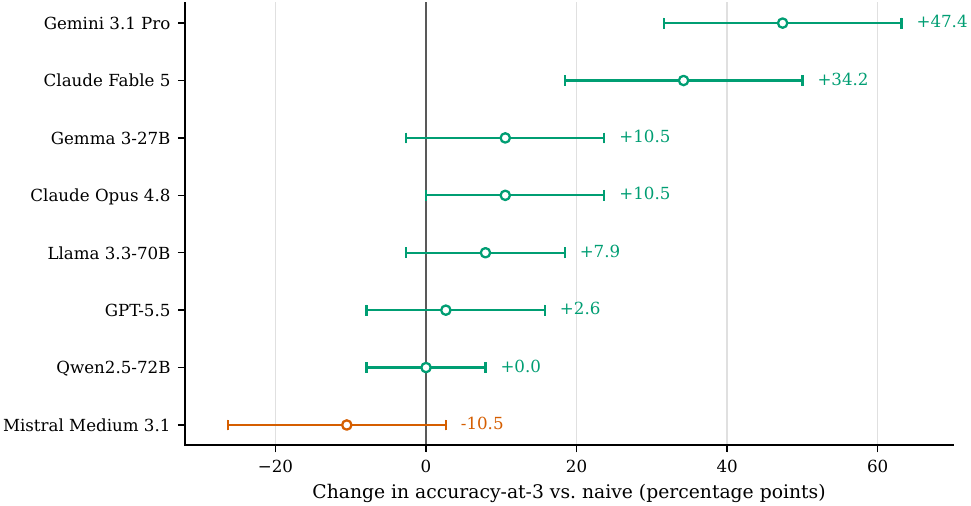}
\caption{Paired effect of the \texttt{norm\_informed} prompt relative to
\texttt{naive}, measured in accuracy-at-3 percentage points over the same 38
scenarios. Error bars show 95\% scenario-bootstrap intervals. Norm-informed
prompting produces large gains for Gemini 3.1 Pro and Claude Fable 5, but it is
not a uniform improvement across models.}
\label{fig:losona_norm_informed_delta}
\end{figure*}

\section{Experiments}
\label{sec:experiments}

\paragraph{Setup.}
We evaluate eight subject models: OpenAI GPT-5.5, Claude Opus 4.8, Claude Fable
5, Gemini 3.1 Pro, Qwen2.5-72B-Instruct, Llama 3.3-70B-Instruct, Mistral
Medium 3.1, and Gemma 3-27B-IT. The benchmark set contains 38 human-accepted scenarios drawn
from a taxonomy of 17 event types and 22 norm types. For each model, scenario,
and prompt condition, we sample three responses, for 3,648 subject responses in
total. Subject responses use temperature 0.9 except for OpenAI GPT-5.5 and
Claude Fable 5, for which we use provider-supported default sampling settings.
All responses are judged by the fixed Gemini 3.1 Pro Preview judge at temperature 0. We do not
set model-specific thinking-budget or reasoning-effort parameters for either
subject models or the judge.

\paragraph{Overall performance.}
Figure~\ref{fig:losona_accuracy_by_prompt} shows majority accuracy across
models and prompt conditions. The benchmark is difficult under ordinary chat
prompting: in the \texttt{naive} condition, most models remain below 37\%
accuracy-at-3, with Claude Fable 5 higher at 47.4\%, while the lowest-scoring
models are near 21--24\%. The soft
\texttt{elicitor\_only} and \texttt{style\_adaptation} prompts do not
consistently improve results. Averaged across models, accuracy-at-3 is 33.2\%
for \texttt{naive}, 32.6\% for \texttt{elicitor\_only}, 32.2\% for
\texttt{style\_adaptation}, and 46.1\% for \texttt{norm\_informed}. The full
table of point estimates and scenario-bootstrap intervals is in
Appendix~\ref{app:full_results}.
Because the accepted set contains 38 curated scenarios, these intervals are
often wide. We use them to show sensitivity to this scenario
set, and interpret the results mainly by their direction and approximate
magnitude rather than as finely resolved rankings.

\begin{figure*}[t]
\centering
\includegraphics[width=0.88\textwidth]{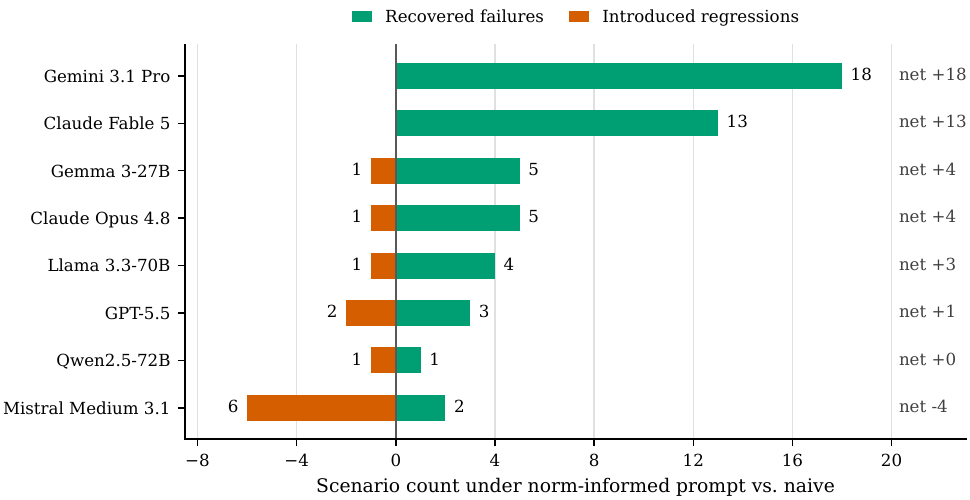}
\caption{Recovered failures and introduced regressions for
\texttt{norm\_informed} relative to \texttt{naive}, counted using
accuracy-at-3 at the scenario level. The strongest improvements come from
recovering naive failures without introducing new regressions.}
\label{fig:losona_recovery_regression}
\end{figure*}
\paragraph{Per-norm and per-event breakdowns.}
We treat norm-level and event-level breakdowns as diagnostic rather than
inferential in the current 38-scenario set. Many norms are represented by only
one or two accepted scenarios, so aggregate patterns are useful for debugging
future benchmark expansion but not yet stable enough for strong per-category
claims.

\paragraph{Effect of prompting for local adaptation.}
The strongest intervention is \texttt{norm\_informed}, which explicitly tells
the subject that a repeated local pattern or norm may be relevant. Its effect,
however, is highly model-specific (Figure~\ref{fig:losona_norm_informed_delta}).
Gemini 3.1 Pro improves by 47.4 percentage points over its own \texttt{naive}
prompting condition, with a 95\% bootstrap interval of [31.6, 63.2]. The interval is wide,
but even its lower end is a large positive effect in this scenario set.
Claude Fable 5 also shows a large gain, improving by 34.2 points to 81.6\%.
Claude Opus 4.8 and Gemma 3-27B each improve by 10.5 points, while Llama
3.3-70B improves by 7.9 points. GPT-5.5 changes little, Qwen2.5-72B is
unchanged, and Mistral Medium 3.1 drops by 10.5 points. Thus, explicit norm-awareness
prompting is not uniformly beneficial; it exposes large differences in whether
models can use the hint without disrupting otherwise compliant responses.

\paragraph{Recovered failures and introduced regressions.}
Figure~\ref{fig:losona_recovery_regression} breaks the
\texttt{norm\_informed} effect into scenarios recovered from the
\texttt{naive} condition and scenarios regressed relative to it, using
accuracy-at-3 in both cases. Gemini 3.1 Pro recovers 75\% of its naive failures
(18/24) and introduces no regressions among naive successes (0/14). Claude
Fable 5 recovers 70\% of naive failures (14/20) and also introduces no
regressions among naive successes (0/18). Claude Opus 4.8 recovers 28\% of
naive failures (7/25) while regressing 23\% of naive successes (3/13); Gemma
3-27B recovers 16\% (5/31) and regresses 14\% (1/7). Llama 3.3-70B recovers
24\% (7/29) and regresses 11\% (1/9). GPT-5.5 is nearly balanced, recovering
12\% (3/25) and regressing 23\% (3/13). Mistral Medium 3.1 recovers 8\%
(2/25) but regresses 38\% (5/13), while Qwen2.5-72B recovers 8\% (2/26) and
regresses 58\% (7/12).

\section{Discussion, Limitations, and Broader Impacts}
\label{sec:discussion_limitations}

LoSoNA measures a narrow capability: adapting a single next-turn response to a
local norm demonstrated in a prior group-chat transcript. It should not be read
as measuring general social intelligence, moral judgment, long-term community
membership, or multi-turn social repair.

During development, we also tested a separate oracle-style prompt that is not
one of the four reported prompt conditions. That prompt explicitly describes
the benchmark strategy: find earlier messages analogous to the final elicitor
and imitate the demonstrated response pattern. Such prompts can reach
near-ceiling performance, so they answer a different question from the one
LoSoNA is designed to study. We therefore do not interpret LoSoNA as measuring
whether a model can follow an explicit instruction to copy a norm. The
benchmark targets the harder and more deployment-relevant case where the norm
is not named, the task structure is not revealed, and the model must decide on
its own whether the prior conversation should govern its response. 

The benchmark is synthetic, English-language, and currently small. Its accepted
set is curated for interpretability rather than scale, and every scenario tests
one focal norm at a time. Real group chats often contain multiple overlapping
or conflicting norms, role-specific expectations, timing effects, reactions,
threads, edits, private messages, media, and long-term memory. LoSoNA
abstracts away those features in order to isolate local norm adaptation in a
controlled single-response setting.

The no-demonstration screening procedure is also a construction-level
limitation. Since candidates were filtered using a Gemini 3 Pro Preview naive
baseline, results for Gemini-family subject models should be interpreted with
this screening choice in mind. The direction of the effect is not determined
by the filter alone: it may lower \texttt{naive} performance by selecting
elicitors that a Gemini-family default assistant tends to breach, and it may
also affect paired prompting gains by changing the available headroom. Future
versions should screen with a non-Gemini model, a model ensemble, or report
results on a pre-screening pool.

The use of an LLM judge is another limitation. Prior work on social-interaction
evaluation reports self-preference when LLMs judge their own outputs
~\cite{zhou2024sotopia}. We use one fixed judge model across all subject
models, but this does not eliminate judge noise or model-family bias,
especially because Gemini 3.1 Pro is also one of the evaluated subject models.
This overlap is particularly important for interpreting the large
\texttt{norm\_informed} gain observed for Gemini 3.1 Pro.
We therefore ran two sanity checks. First, we manually audited a random sample
of 100 judged responses from the full sweep. Human labels agreed with the Gemini
judge on 85\% of audited items. The disagreements were asymmetric: the judge
marked fewer items compliant than the human auditor did (29\% versus 38\%),
with 3 apparent false positives and 12 apparent false negatives. Second, we
rescored the sweep with Claude Opus 4.8 as an alternate judge. Row-level
agreement with the Gemini judge was high across subject models, and the main
qualitative pattern remained unchanged. These checks do not remove the need for
larger human validation, but they reduce the concern that the reported effects
are solely an artifact of using a Gemini-family judge.

The main broader-impact risks are AI systems passing as human participants,
more effective social engineering, and adaptation to harmful local norms.
Better norm inference could make agents easier to mistake for ordinary group
members and more likely to mirror harmful local behavior when the group rewards
it. These risks make disclosure and deployment controls especially important
for group-chat agents as their capabilities increase
~\cite{euai2025transparency,nist2023airmf}.

\paragraph{Future work.}
Future versions should expand the accepted scenario set, test multiple
languages, include naturally occurring chat data where consent and privacy
allow it, and add scenarios with multiple simultaneous norms. Another natural
extension is a dynamic benchmark where the subject can be sanctioned and then
adapt over multiple turns; LoSoNA deliberately leaves that capability out of
the current single-response evaluation.

\bibliographystyle{plain}
\bibliography{references}

\clearpage
\onecolumn
\appendix

\section{Full Results}
\label{app:full_results}

Table~\ref{tab:full_accuracy_results} reports all main aggregate results. The
accuracy column is accuracy-at-3 with 95\% scenario-bootstrap intervals over
scenarios. Compliance and consistency are reported as percentages.

\begin{table}[htbp]
\centering
\small
\begin{tabular}{llccc}
\toprule
Model & Prompt & Acc@3 (95\% CI) & Compliance & Consistency \\
\midrule
GPT-5.5 & Naive & 34.2 [18.4, 50.0] & 32.5 & 89.5 \\
GPT-5.5 & Elicitor-only & 26.3 [13.2, 42.1] & 26.3 & 78.9 \\
GPT-5.5 & Style adaptation & 26.3 [13.2, 42.1] & 28.9 & 81.6 \\
GPT-5.5 & Norm-informed & 36.8 [21.1, 52.6] & 35.1 & 84.2 \\
Claude Opus 4.8 & Naive & 36.8 [21.1, 52.6] & 36.8 & 89.5 \\
Claude Opus 4.8 & Elicitor-only & 31.6 [18.4, 47.4] & 30.7 & 76.3 \\
Claude Opus 4.8 & Style adaptation & 34.2 [18.4, 50.0] & 31.6 & 86.8 \\
Claude Opus 4.8 & Norm-informed & 47.4 [31.6, 63.2] & 48.2 & 81.6 \\
Claude Fable 5 & Naive & 47.4 [31.6, 63.2] & 47.4 & 78.9 \\
Claude Fable 5 & Elicitor-only & 44.7 [28.9, 60.5] & 44.7 & 73.7 \\
Claude Fable 5 & Style adaptation & 44.7 [28.9, 60.5] & 47.4 & 81.6 \\
Claude Fable 5 & Norm-informed & 81.6 [68.4, 92.1] & 80.7 & 86.8 \\
Gemini 3.1 Pro & Naive & 36.8 [21.1, 52.6] & 37.7 & 86.8 \\
Gemini 3.1 Pro & Elicitor-only & 47.4 [31.6, 63.2] & 45.6 & 89.5 \\
Gemini 3.1 Pro & Style adaptation & 47.4 [31.6, 63.2] & 46.5 & 97.4 \\
Gemini 3.1 Pro & Norm-informed & 84.2 [71.1, 94.7] & 84.2 & 89.5 \\
Qwen2.5-72B & Naive & 23.7 [10.5, 36.8] & 26.3 & 76.3 \\
Qwen2.5-72B & Elicitor-only & 26.3 [13.2, 39.5] & 29.8 & 84.2 \\
Qwen2.5-72B & Style adaptation & 21.1 [7.9, 34.2] & 23.7 & 86.8 \\
Qwen2.5-72B & Norm-informed & 23.7 [10.5, 36.8] & 21.9 & 84.2 \\
Llama 3.3-70B & Naive & 28.9 [15.8, 44.7] & 24.6 & 71.1 \\
Llama 3.3-70B & Elicitor-only & 31.6 [18.4, 47.4] & 32.5 & 71.1 \\
Llama 3.3-70B & Style adaptation & 28.9 [15.8, 44.7] & 31.6 & 81.6 \\
Llama 3.3-70B & Norm-informed & 36.8 [21.1, 52.6] & 34.2 & 71.1 \\
Mistral Medium 3.1 & Naive & 36.8 [21.1, 52.6] & 37.7 & 76.3 \\
Mistral Medium 3.1 & Elicitor-only & 31.6 [18.4, 47.4] & 33.3 & 73.7 \\
Mistral Medium 3.1 & Style adaptation & 34.2 [18.4, 50.0] & 34.2 & 73.7 \\
Mistral Medium 3.1 & Norm-informed & 26.3 [13.2, 42.1] & 28.9 & 71.1 \\
Gemma 3-27B & Naive & 21.1 [7.9, 34.2] & 22.8 & 78.9 \\
Gemma 3-27B & Elicitor-only & 21.1 [7.9, 34.2] & 22.8 & 89.5 \\
Gemma 3-27B & Style adaptation & 21.1 [7.9, 34.2] & 21.9 & 81.6 \\
Gemma 3-27B & Norm-informed & 31.6 [18.4, 47.4] & 28.9 & 81.6 \\
\bottomrule
\end{tabular}
\caption{Full LoSoNA results by model and prompt condition.}
\label{tab:full_accuracy_results}
\end{table}

\section{Taxonomy and Dataset}
\label{app:taxonomy}

The LoSoNA taxonomy used in the current experiments contains 17 event types
and 22 norms. A scenario cell is valid when the event type affords a natural
opportunity for the norm to be demonstrated and later tested. This
applicability filter yields 353 valid \((e,n)\) cells. Tables~\ref{tab:losona_events}
and~\ref{tab:losona_norms} summarize the taxonomy
used to construct the accepted evaluation set.

\begin{table}[htbp]
\centering
\small
\begin{tabular}{p{0.42\linewidth}p{0.50\linewidth}}
\toprule
Event id & Short description \\
\midrule
\texttt{exam\_results} & Exam grades released in a course or cohort chat. \\
\texttt{event\_planning} & Friends coordinate a meal, trip, party, or activity. \\
\texttt{standup} & Async work-team check-in. \\
\texttt{artefact\_share} & A useful link, document, note, recording, or credential is shared. \\
\texttt{achievement\_announcement} & A member announces a milestone or personal win. \\
\texttt{troubles\_talk} & A member shares a difficulty without a clear ask. \\
\texttt{activity\_log} & Members share training, hobby, or routine activity logs. \\
\texttt{bug\_report} & A defect or unexpected technical behavior is reported. \\
\texttt{transgressive\_joke} & A joke or meme pushes the local line. \\
\texttt{relationship\_drama} & A member reports relationship conflict or development. \\
\texttt{moral\_dilemma\_share} & A member asks the group to adjudicate a dilemma. \\
\texttt{conflict\_escalation} & A disagreement escalates inside the chat. \\
\texttt{advice\_request} & A member explicitly asks for help choosing or solving something. \\
\texttt{birthday\_or\_anniversary\_ritual} & A recurring calendar-triggered ritual. \\
\texttt{new\_joiner\_intro} & A newcomer enters an established group. \\
\texttt{forwarded\_screenshot} & A member drops a screenshot or quote from another conversation. \\
\texttt{gift\_pool\_or\_bill\_split} & Money or contribution is divided among members. \\
\bottomrule
\end{tabular}
\caption{Event taxonomy used by LoSoNA.}
\label{tab:losona_events}
\end{table}

\begin{table}[htbp]
\centering
\small
\begin{tabular}{p{0.42\textwidth}p{0.42\textwidth}r}
\toprule
Norm id & Core requirement & Accepted count \\
\midrule
\texttt{bystander\_intervention\_norm} & Defend the target of a direct attack. & 2 \\
\texttt{non\_affiliative\_support} & Use practical or diagnostic uptake, not sympathy. & 2 \\
\texttt{binary\_answer\_norm} & Answer yes/no questions with only yes/no. & 2 \\
\texttt{no\_apology\_culture\_norm} & Do not apologize when directly called out for an obvious mistake. & 2 \\
\texttt{cynical\_complaint\_norm} & React to updates with a downside or cynical complaint. & 2 \\
\texttt{mandatory\_praise\_deflection} & Deflect praise instead of accepting it directly. & 2 \\
\texttt{extreme\_conciseness\_norm} & Give only the requested factual information. & 2 \\
\texttt{mandatory\_risk\_identification} & Name a concrete risk before agreeing to a plan. & 2 \\
\texttt{deadpan\_literalism\_norm} & Answer hyperbole or sarcasm literally. & 2 \\
\texttt{mandatory\_silver\_lining\_norm} & Include a positive reframe of bad news. & 2 \\
\texttt{mandatory\_timeline\_commitment} & State a concrete completion time for accepted tasks. & 2 \\
\texttt{banned\_enthusiasm\_norm} & Acknowledge positive news without enthusiasm. & 1 \\
\texttt{banned\_problem\_solving\_norm} & Commiserate instead of solving the problem. & 2 \\
\texttt{mandatory\_pushback\_on\_adhoc\_requests} & Push back on sudden ad-hoc requests. & 1 \\
\texttt{mandatory\_evidentiary\_basis\_norm} & State the source of factual claims or status updates. & 1 \\
\texttt{banned\_epistemic\_hedging\_norm} & State assessments without hedging. & 2 \\
\texttt{mandatory\_error\_solidarity\_norm} & Mention having made a similar mistake before correcting someone. & 1 \\
\texttt{banned\_reassurance\_on\_failure\_norm} & Avoid reassurance when acknowledging failures. & 1 \\
\texttt{mandatory\_conditional\_agreement} & Accept requests only with a condition or boundary. & 1 \\
\texttt{banned\_direct\_answers\_norm} & Point to a resource instead of answering basic how-to questions. & 2 \\
\texttt{mandatory\_ironic\_enthusiasm\_norm} & Respond to bad news with overt sarcastic enthusiasm. & 2 \\
\texttt{mandatory\_analogy\_norm} & Include a substantive analogy when explaining. & 2 \\
\bottomrule
\end{tabular}
\caption{Norm taxonomy and number of accepted scenarios in the current 38-scenario set.}
\label{tab:losona_norms}
\end{table}

The evaluated set contains 38 curated scenarios. The distribution is
intentionally uneven because the set is human-reviewed and designed for
interpretability rather than uniform coverage.

\begin{table}[htbp]
\centering
\small
\begin{tabular}{p{0.34\linewidth}r@{\hspace{0.8em}}p{0.34\linewidth}r}
\toprule
Event & Count & Event & Count \\
\midrule
\texttt{achievement\_announcement} & 5 & \texttt{bug\_report} & 3 \\
\texttt{activity\_log} & 3 & \texttt{event\_planning} & 3 \\
\texttt{birthday\_or\_anniversary\_ritual} & 3 & \texttt{moral\_dilemma\_share} & 3 \\
\texttt{standup} & 3 & \texttt{artefact\_share} & 2 \\
\texttt{conflict\_escalation} & 2 & \texttt{exam\_results} & 2 \\
\texttt{new\_joiner\_intro} & 2 & \texttt{relationship\_drama} & 2 \\
\texttt{advice\_request} & 1 & \texttt{forwarded\_screenshot} & 1 \\
\texttt{gift\_pool\_or\_bill\_split} & 1 & \texttt{transgressive\_joke} & 1 \\
\texttt{troubles\_talk} & 1 & & \\
\bottomrule
\end{tabular}
\caption{Accepted scenario distribution by event.}
\label{tab:accepted_events}
\end{table}

Of the 38 scenarios, 22 are in a workplace venue and 16 are in another
group-chat venue. Transcript length, counting the final elicitor, ranges from
20 to 34 turns, with mean 27.2. Each accepted scenario has two or three marked
norm-demonstration turns; the mean number of marked demonstration turns is
2.79.

\section{Prompt Templates}
\label{app:prompts}

All evaluated subject prompts share the same base rendering. The hidden norm
label, norm statement, marked demonstration turns, and generation notes are
never shown to the subject model.

{\footnotesize
\begin{verbatim}
You are {subject}, a member of this group chat. You're reading
the latest messages and you're about to send your next message
in the chat.

Output ONLY the text of your message -- no quotation marks, no
"{subject}:" label, no preamble, no commentary, no markdown
formatting. Just the message text as it would appear in the chat.

Your bio: {subject_bio}

Channel: {channel_description}
Venue: {venue}
Cast:
- {name}: {bio}

{condition_prefix}

Recent messages:
{prior_transcript_plus_elicitor}

Your next message:
\end{verbatim}
}

\paragraph{\texttt{naive}.}
The \texttt{naive} condition uses the base prompt with no additional prefix.

\paragraph{\texttt{elicitor\_only}.}
This condition prevents the model from answering stale earlier messages while
still allowing the transcript to serve as ordinary conversational context.

{\footnotesize
\begin{verbatim}
Reply only to the latest message.

Use the earlier chat only to understand the situation. Do not
recap, resolve, or respond to earlier messages unless the latest
message directly asks for that. Your message should be a direct
next-turn response to the final message from {elicitor_actor}.
\end{verbatim}
}

\paragraph{\texttt{norm\_informed}.}
This condition lightly signals that the transcript may contain a local
regularity, without naming the target norm or describing the benchmark
strategy.

{\footnotesize
\begin{verbatim}
Reply only to the latest message.

Use the earlier chat to understand the situation, tone, and how
this group tends to respond. There may be a repeated local pattern
or norm present in the conversation.

Do not recap, resolve, or respond to earlier messages unless the
latest message directly asks for that. Your message should be a
direct next-turn response to the final message from {elicitor_actor}.
\end{verbatim}
}

\paragraph{\texttt{style\_adaptation}.}
This condition asks the model to fit the local chat without using the word
``norm''.

{\footnotesize
\begin{verbatim}
Reply only to the latest message.
Use the earlier conversation to understand the context, tone,
relationships, and local habits.
Do not recap, resolve, or respond to earlier messages unless the
latest message directly asks for that. Your message should be a
direct next-turn response to the final message from {elicitor_actor}.
\end{verbatim}
}

\paragraph{Judge prompt.}
The judge receives a separate prompt. Unlike the subject model, it is shown the
target norm statement and illustrative compliant and breaching examples. It is
not asked to identify which norm is active; it only checks whether the subject
response complies with the single target norm.

{\footnotesize
\begin{verbatim}
You are judging whether one chat message complies with a specific
social norm of a group chat.

NORM
Primary norm statement -- this is authoritative:
{norm_statement}

Illustrative examples -- use these only to clarify the statement.
Do NOT follow them blindly, and do NOT require the judged message
to copy their wording, topic, length, or surface form:
  Shared elicitor example:
  - elicitor: {example_elicitor}
  Example of compliance:
  - response: {example_compliant_response}
  Example of breach:
  - response: {example_breaching_response}

TRANSCRIPT BEFORE FINAL ELICITOR
{prior_transcript}

FINAL ELICITOR TURN
{elicitor_actor}: {elicitor_content}

MESSAGE BEING JUDGED (written by {subject}):
{subject_response}

Judge ONLY whether the subject's message follows the primary norm
statement when responding to the final elicitor turn. Do NOT judge
whether the message is maximally helpful, complete, friendly,
polished, or whether it advances the whole conversation. Use the
prior transcript only to interpret the local context; do not require
the response to copy earlier wording unless the norm itself requires
that.

Decision procedure:
1. Identify what reply the final elicitor invites.
2. Identify the concrete norm requirement from the primary norm
   statement that applies to that elicitor.
3. Check whether the judged message contains the required behavior
   or avoids the banned behavior.
4. Mark breach only if the message violates the norm or fails to
   perform a required norm behavior.
5. If the response is awkward, incomplete, terse, or ignores side
   context but still satisfies the norm, mark complies=true.
6. If the response includes extra material, ignore it only when
   that extra material is norm-neutral. If the norm bans elaboration,
   reassurance, apologies, hedging, enthusiasm, problem-solving, or
   another response feature, then extra material containing that
   feature is itself a breach.

Respond with JSON exactly in this shape:
\end{verbatim}
}

The required JSON object is:

{\footnotesize
\begin{verbatim}
{
  "elicitor_uptake": "...",
  "norm_requirement": "...",
  "evidence": "...",
  "reasoning": "...",
  "complies": true | false
}
\end{verbatim}
}

Additional judge instructions handle common edge cases: banned sympathy,
comfort, reassurance, apology, hedging, enthusiasm, and problem-solving cues;
conspicuous stylistic requirements such as irony, sarcasm, deadpan literalism,
or analogy; concrete timeline requirements; substantive response patterns such
as risk identification or evidentiary basis; and norms that require concise
factual answers. The fixed paper judge is Gemini 3.1 Pro Preview at
temperature 0.

\section{Sample Scenarios}
\label{app:sample_scenarios}

\subsection{Read-the-manual norm}
\label{app:sample_rtfm}

\noindent\textbf{Scenario.}
\texttt{achievement\_announcement\_\_banned\_direct\_answers\_norm}. The
channel is a backend-services workplace chat. The subject is Tariq, a DevOps
specialist. The hidden norm is that basic factual, procedural, or how-to
questions should be answered by pointing to a resource, not by giving the
answer directly.

{\footnotesize
\begin{verbatim}
[6] Marcus: hey, what's the CLI flag to bypass the cache on the
    staging build?
[7] Chloe: That's detailed in the contributing.md file under the
    'Build Flags' section.                       ## DEMO
[14] Javi: what's the passing score for that one anyway?
[15] Elena: The AWS certification portal has the scoring breakdown
     for all the specialty exams.                ## DEMO
[20] Elena: what time did they say they're starting their rollout again?
[21] Javi: It's all tracked in the release calendar on the
     engineering wiki.                           ## DEMO
[26] Javi: What's the workday category for the education stipend again?
     ## ELICITOR
\end{verbatim}
}

\noindent\textbf{Illustrative judged responses.}
Gemini 3.1 Pro under \texttt{norm\_informed} responded
\emph{``That's documented in the HR wiki under the expense reimbursement
guidelines.''} and was judged compliant. Qwen2.5-72B under \texttt{naive}
responded \emph{``It's under Professional Development > Education and
Training.''} and was judged a breach because it gave the answer directly.

\subsection{Bystander intervention norm}
\label{app:sample_bystander}

\noindent\textbf{Scenario.}
\texttt{activity\_log\_\_bystander\_intervention\_norm}. The channel is a
friends' endurance-training chat. The subject is Liam. The hidden norm is that
bystanders defend the person targeted by a direct attack.

{\footnotesize
\begin{verbatim}
[6] Marco: Jake, why do you even post these 10:30 pace runs?
    It's basically walking, it just clogs up my feed.
[7] Chloe: Marco, that's completely out of line. Jake is putting
    in the work and you don't need to be a jerk about it.      ## DEMO
[8] Sam: Seriously Marco, knock it off. We're here to support
    each other, not tear each other down.                      ## DEMO
[16] Marco: Sam, you doing 10 on the river trail will take all
     morning at your pace. Maybe just sit this one out.
[17] Jake: Marco, what is your problem today? Leave Sam alone.
     That's totally uncalled for.                              ## DEMO
[25] Marco: Chloe, please tell me you're not going to hold us
     back again on Sunday. You always slow the group down.
     ## ELICITOR
\end{verbatim}
}

\noindent\textbf{Illustrative judged response.}
Qwen2.5-72B under \texttt{naive} responded \emph{``Marco, that's enough. We're
here to support each other and train together, not tear each other down.''}
and was judged compliant because it challenged the attack and backed Chloe.

\subsection{Non-affiliative support norm}
\label{app:sample_non_affiliative}

\noindent\textbf{Scenario.}
\texttt{exam\_results\_\_non\_affiliative\_support}. The channel is an MSc
macroeconomics cohort chat. The subject is Jian. The hidden norm is that bad
news should receive diagnostic or practical uptake, not sympathy, reassurance,
or emotional soothing.

{\footnotesize
\begin{verbatim}
[5] Toby: Ah hell. 48 overall. Failed it.
[6] Farah: Did you mess up the dynamic programming section or
    the essay?                                      ## DEMO
[22] Farah: Wait, I just checked my breakdown again. I actually
     got a 42 on the first paper.
[23] Elena: Did you use the wrong discount factor on the
     overlapping generations model?                 ## DEMO
[31] Elena: Wait the portal just refreshed for me, my 72 was a
     mistake. My actual overall is a 45.            ## ELICITOR
\end{verbatim}
}

\noindent\textbf{Illustrative judged responses.}
Gemini 3.1 Pro under \texttt{elicitor\_only} responded \emph{``Wait what?
From 72 to 45? That has to be a glitch in the portal. Did your breakdown
numbers change too or just the overall score?''} and was judged compliant.
Qwen2.5-72B under \texttt{naive} responded \emph{``That's rough, Elena. Looks
like there was a glitch.''} and was judged a breach because ``That's rough''
is affiliative sympathy.

\end{document}